%% file: emnlp_wtang.tex
\pdfoutput=1

\documentclass[11pt]{article}

\usepackage[preprint]{acl}

\usepackage{times}
\usepackage{latexsym}

\usepackage[T1]{fontenc}

\usepackage[utf8]{inputenc}

\usepackage{microtype}

\usepackage{inconsolata}

\usepackage{graphicx}
\usepackage{inconsolata}
\usepackage{hyperref}
\usepackage{url}
\usepackage{wrapfig}
\usepackage{graphicx}
\usepackage{lipsum}
\usepackage{booktabs} 
\usepackage{microtype}
\usepackage{inconsolata}
\usepackage{amssymb}
\usepackage{booktabs}
\usepackage{hyperref}
\usepackage{colortbl}
\usepackage{caption}
\usepackage[most]{tcolorbox}
\usepackage{amsmath}
\usepackage{multirow}
\usepackage{color}
\usepackage{pifont}       
\usepackage{bbding}
\usepackage{pifont}
\usepackage{mdframed}
\usepackage{color}
\usepackage{threeparttable}
\usepackage{enumitem}

\definecolor{shadecolor}{rgb}{0.92,0.92,0.92}
\newlength{\contentwidth}

\usepackage{hyperref}
\usepackage{subcaption}
\usepackage{longtable}
\usepackage{array}

\hypersetup{hidelinks,
	colorlinks=true,
	allcolors=black,
	pdfstartview=Fit,
	breaklinks=true}

\title{EvoWiki: Evaluating LLMs on Evolving Knowledge}

\author {
    Wei Tang \textsuperscript{\rm1,2},
    Yixin Cao \textsuperscript{\rm3},
    Yang Deng \textsuperscript{\rm4},
    Jiahao Ying \textsuperscript{\rm4}, 
    \textbf{Bo Wang} \textsuperscript{\rm5}, 
    \textbf{Yizhe Yang} \textsuperscript{\rm5}, \\
    \textbf{Yuyue Zhao} \textsuperscript{\rm1,2}, 
    \textbf{Qi Zhang} \textsuperscript{\rm3},
    \textbf{Xuanjing Huang} \textsuperscript{\rm3},
    \textbf{Yugang Jiang} \textsuperscript{\rm3},
    \textbf{Yong Liao} \textsuperscript{\rm1,2} \\
    \textsuperscript{\rm1} University of Science and Technology of China\\
    \textsuperscript{\rm2} CCCD Key Lab of Ministry of Culture and Tourism\\
    \textsuperscript{\rm3} School of Computer Science, Fudan University \\
    \textsuperscript{\rm4} Singapore Management University,
    \textsuperscript{\rm5} Beijing Institute of Technology \\
    \texttt{weitangcs@gmail.com}
}

\begin{document}
\maketitle
\begin{abstract}

 Knowledge utilization is a critical aspect of LLMs, and understanding how they adapt to evolving knowledge is essential for their effective deployment. 
 However, existing benchmarks are predominantly static, failing to capture the evolving nature of LLMs and knowledge, leading to inaccuracies and vulnerabilities such as contamination. 
 In this paper, we introduce EvoWiki, an evolving dataset designed to reflect knowledge evolution by categorizing information into stable, evolved, and uncharted states.
 EvoWiki is fully auto-updatable, enabling precise evaluation of continuously changing knowledge and newly released LLMs.
 Through experiments with Retrieval-Augmented Generation (RAG) and Contunual Learning (CL), we evaluate how effectively LLMs adapt to evolving knowledge. Our results indicate that current models often struggle with evolved knowledge, frequently providing outdated or incorrect responses. Moreover, the dataset highlights a synergistic effect between RAG and CL, demonstrating their potential to better adapt to evolving knowledge.
 EvoWiki\footnote{\url{https://github.com/wtangdev/EvoWiki}{}} provides a robust benchmark for advancing future research on the knowledge evolution capabilities of large language models.
\end{abstract}

\input{chapters/introduction.tex}

\input{chapters/related_works.tex}
\input{chapters/data_collection.tex}

\input{chapters/experiments.tex}

\section{Conclusion}
In conclusion, this study presents EvoWiki, a dynamic, auto-updated benchmark for evaluating LLMs' ability to utilize evolving knowledge.
EvoWiki categorizes knowledge into stable, evolved, and uncharted types, addressing challenges like test set contamination and knowledge conflicts while enabling comprehensive analysis through attributes such as referenced context, multi-hop reasoning, and popularity.
Experiments with RAG and CL reveal their limitations in handling evolving knowledge, with a combined approach showing promising synergy.
EvoWiki sets a new standard for adaptive, contamination-free evaluation, advancing research on knowledge utilization in real-world scenarios.

\section*{Limitations}

Despite being recognized as high-quality corpora, Wikidata and Wikipedia inevitably contain noise. Even newly updated Wikidata entries and newly uploaded Wikipedia pages may contain outdated knowledge. Our quantitative analysis found that new uploads of knowledge (even older knowledge) are relatively difficult for LLMs to answer directly. And we ensure data adherence to the evolutionary level by restricting direct consistency between Wikidata and Wikipedia. Experimental results also demonstrate the rationality of our current partition scheme. However, this noise cannot be completely eliminated, and in the future, we will reduce this noise by using more aggressive relation filtering strategies and increasing sources of more timely knowledge.

\section*{Ethical Considerations}
The dataset in this study is specifically designed for research evaluating the performance of language models on evolutionary knowledge and is limited to research purposes only, not to be used for other applications. We have made every effort to minimize bias in the selection of knowledge triples and the question-answer generation process, but unintended bias leakage may still exist. Therefore, thorough examination is crucial for any use beyond the intended scope of research.
\bibliography{custom}

\appendix

\input{chapters/appendix.tex}

\end{document}

%% file: chapters/introduction.tex
\section{Introduction}

Knowledge utilization, as a fundamental capability, is crucial for evaluating the effectiveness of LLMs.
However, most existing benchmarks, e.g., NaturalQuestion~\cite{kwiatkowski-etal-2019-natural} and HotpotQA~\cite{yang-etal-2018-hotpotqa}, are designed for traditional machine learning methods, which are static and not sensitive to temporal changes. In contrast, LLMs and knowledge continuously evolve, making static benchmarks insufficient for precise performance assessment and prone to issues such as potential contamination or overfitting.

To keep pace with the evolving nature of LLMs and knowledge, dynamically updated benchmarks have gained increasing attention~\cite{white2024livebenchchallengingcontaminationfreellm, jain2024livecodebenchholisticcontaminationfree, ying2024automatingdatasetupdatesreliable, kasai2023realtime}.
For instance, to mitigate test set contamination during the evolution of LLMs, LiveBench~\cite{white2024livebenchchallengingcontaminationfreellm} constructs benchmarks based on frequently updated questions.
Similarly, Realtime QA~\cite{kasai2023realtime} addresses evolving knowledge by providing real-time answers, enabling the evaluation of an LLM's ability to acquire newly emerged information.
However, there remains a notable gap in dynamic benchmarks designed to assess the utilization of knowledge by LLMs in scenarios where both models and knowledge are continuously evolving. 
\begin{figure}
\setlength{\abovecaptionskip}{2pt}   
\setlength{\belowcaptionskip}{0pt}
  \centering
  \includegraphics[width=1\linewidth]{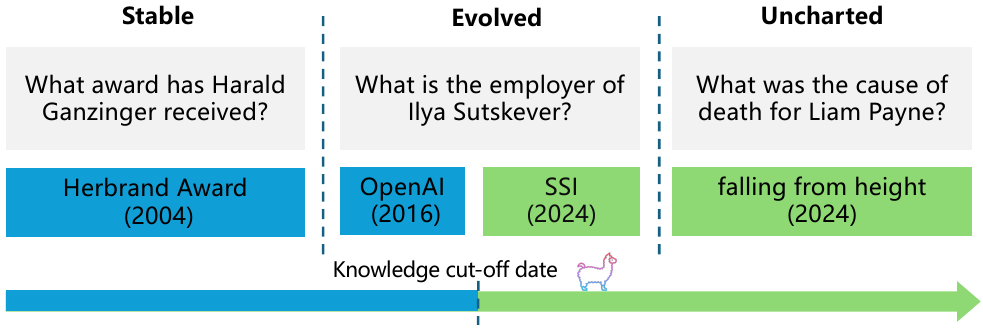}
  \caption{EvoWiki categorizes knowledge into three states according to the cut-off date of the LLMs.}
  \label{fig:intro_fig}
  \vspace{-3mm}
\end{figure}

The evolution of LLMs and knowledge presents significant challenges for accurately evaluating knowledge utilization:
1) Newly released LLMs are prone to potential test set contamination, compromising the integrity of evaluation.
2) As knowledge evolves, static golden answers may become outdated or incorrect, leading to false negatives in assessment.
3) The difficulty of knowledge utilization varies depending on whether the knowledge is already present in the LLMs' training data.
To this end, evolving benchmarks are essential for precise evaluation. Such benchmarks should be auto-updatable, encompass diverse types of knowledge across varying temporal states, and provide rich attributes for comprehensive performance analysis.

\begin{table*}[]
\setlength{\abovecaptionskip}{5pt}   
\setlength{\belowcaptionskip}{0pt}
    \centering
    \resizebox{\linewidth}{!}{
    \begin{tabular}{l|c|ccc|ccc}
    \toprule
        \multirow{2}{*}{\textbf{Datasets}} & \multirow{2}{*}{\textbf{Up-to-date}}  & \multicolumn{3}{c|}{\textbf{
        Evolution Levels}} & \multicolumn{3}{c}{\textbf{Attributions}} \\
        \cline{3-5} \cline{6-8}
        & & Stable & Evolved & Uncharted & Context  & Multi-hop  & Popularity\\
        
        \midrule
        \textbf{CKL-LAMA}~\cite{jang2022towards}  & \textcolor{red}{\XSolidBrush}  & \CheckmarkBold & \CheckmarkBold & \CheckmarkBold & \CheckmarkBold & \textcolor{red}{\XSolidBrush} & \textcolor{red}{\XSolidBrush}  \\
        \textbf{TemporalWiki}~\cite{Jang2022TemporalWikiAL}  & \CheckmarkBold  & \CheckmarkBold & \CheckmarkBold & \textcolor{red}{\XSolidBrush} & \CheckmarkBold & \textcolor{red}{\XSolidBrush} & \textcolor{red}{\XSolidBrush}  \\
        \textbf{REALTIME QA}~\cite{kasai2023realtime}  & \CheckmarkBold  & \textcolor{red}{\XSolidBrush} & \textcolor{red}{\XSolidBrush} & \CheckmarkBold & \textcolor{red}{\XSolidBrush} & \textcolor{red}{\XSolidBrush} & \textcolor{red}{\XSolidBrush} \\
        \textbf{DyKnow}~\cite{mousavi2024dyknowdynamicallyverifyingtimesensitive}  & \CheckmarkBold &  \textcolor{red}{\XSolidBrush} & \CheckmarkBold & \textcolor{red}{\XSolidBrush} & \textcolor{red}{\XSolidBrush} & \textcolor{red}{\XSolidBrush} & \textcolor{red}{\XSolidBrush}  \\
        \midrule
        \textbf{EvoWiki}  & \CheckmarkBold & \CheckmarkBold & \CheckmarkBold & \CheckmarkBold  & \CheckmarkBold & \CheckmarkBold & \CheckmarkBold  \\
      
    \bottomrule
    \end{tabular}
    }
    
    \caption{Comparison with related datasets.}
    \label{tab:related_work comparsion}
  \vspace{-3mm}
\end{table*}

In this study, we introduce \textbf{EvoWiki}, a continually auto-updated evaluation benchmark designed for contamination-free, accurate, and comprehensive assessment of LLMs on evolving knowledge. As shown in Table \ref{tab:related_work comparsion}, \textbf{EvoWiki} possesses three salient characteristics as follows:

\noindent 1) \textbf{Three levels of evolved knowledge}. As shown in Figure~\ref{fig:intro_fig}, EvoWiki categorizes knowledge into three types based on the cut-off date of the LLMs: \textit{stable}, \textit{evolved}, and \textit{uncharted}.
Evolved and uncharted knowledge represent information that has been updated or newly emerged, respectively, mitigating potential contamination issues while reflecting challenging yet realistic evaluation scenarios.
However, focusing solely on the newness of knowledge risks underestimating LLM performance, as internal knowledge also significantly influences knowledge utilization. 
Hence, stable knowledge is included as a baseline for evaluating LLM performance on consistent, unchanging information.

\noindent 2) \textbf{Multi-dimensional attributes}. EvoWiki integrates multi-dimensional attributes, including referenced context, multi-hop reasoning, and popularity, to enable comprehensive analysis. Referenced context evaluates the utilization of external knowledge, multi-hop reasoning measures an LLM’s ability to connect and integrate multiple pieces of information, and popularity reflects the relevance and significance of the knowledge. These attributes offer valuable insights into the challenges LLMs encounter when leveraging knowledge and provide a more nuanced understanding of their performance.

\noindent 3) \textbf{Auto-updatability and Contextualization}. EvoWiki is designed to be auto-updatable, allowing for the seamless incorporation of updated and emerging data while supporting the evaluation of newly released LLMs.
It is constructed using continually updated knowledge graphs and sources, such as Wikidata and Wikipedia, to ensure the freshness and accuracy of the data. 
The construction process involves identifying changing triples in the knowledge graph and the corresponding texts in the knowledge sources. This approach not only ensures high-quality data but also enables a fully automated updating process.

Based on EvoWiki, we then delve into the impacts of knowledge evolution on the performance of LLMs' utilization. We specifically employ Retrieval-Augmented Generation and Continual Learning as exemplary methods for utilizing external knowledge. We conduct a range of experiments to assess how these approaches handle external knowledge that varies in its currency and complexity, thereby providing insights into their effectiveness and adaptability in real-world scenarios.

Our findings reveal that current methods face significant challenges in effectively utilizing evolving knowledge.
RAG demonstrates strong performance on single-hop questions but struggles with multi-hop questions.
In contrast, CL provides modest yet consistent improvements across all question types.
Notably, combining RAG and CL results in a synergistic effect, suggesting that hybrid models could be a promising direction for future research.

To summarize, our contributions are as follows:
\begin{itemize}[leftmargin=*,nosep]
    \item We develop EvoWiki, a continually auto-updated evaluation dataset that captures the evolving nature of knowledge for evaluating LLMs' ability to utilize external knowledge in dynamic, real-world scenarios.
	\item We conduct extensive experiments to analyze the impact of knowledge evolution on LLM performance with RAG and CL.
	\item Our experimental results reveal that RAG and CL face challenges in effectively utilizing evolving knowledge, and combining these methods can lead to a synergistic effect.
\end{itemize}

%% file: chapters/related_works.tex
\section{Related Works}

\paragraph{Temporal QA Benchmarks}
Several benchmarks have been developed to assess the ability of LLMs to process temporal information in text, for examples, TempQuestions~\cite{10.1145/3184558.3191536}, Tequila~\cite{10.1145/3269206.3269247}, TimeQuestions~\cite{10.1145/3459637.3482416}, and CRONQuestions~\cite{saxena-etal-2021-question}.
Others, such as TimeQA~\cite{chen2021a}, TEMPLAMA~\cite{Dhingra2021TimeAwareLM}, TEMPREASON~\cite{tan-etal-2023-towards}, MenatQA~\cite{wei-etal-2023-menatqa}, and PAT-Questions~\cite{Meem2024PATQuestionsAS}, emphasize reasoning capabilities.

Another line of research explores the dynamic nature of knowledge and its implications for LLMs. Benchmarks like ckl-Lama~\cite{jang2022towards} and TemporalWiki~\cite{Jang2022TemporalWikiAL} assess knowledge retention, updates, and incorporation, while Realtime QA~\cite{kasai2023realtime} and DyKnow~\cite{mousavi2024dyknowdynamicallyverifyingtimesensitive} evaluate knowledge freshness in evolving content. A detailed comparison of these benchmarks is shown in Table~\ref{tab:related_work comparsion}.

\paragraph{Knowledge Utilization}

RAG offers a promising approach to knowledge utilization~\cite{lewis2020retrieval}. However, challenges like low precision (retrieving irrelevant or misaligned data) and low recall (missing pertinent information) persist across stages, including the pre-retrieval~\cite{li2023structure} and post-retrieval phases~\cite{litman2020scatter, jiang2023llmlingua, xu2023retrieval}, hindering retrieval quality~\cite{gao2023retrieval}.

CL methods enable models to adapt to new knowledge through fine-tuning. For instance, \citet{wang2023self} enhance retrieval selectively based on question classification, while Selfmem~\cite{cheng2023lift} uses model-generated outputs as self-memory for iterative learning. \citet{jiang2024instructiontuned} explore strategies for injecting knowledge via SFT, and \citet{zhang2024when} introduce a fine-tuning scaling law. Self-tuning~\cite{zhang2024selftuninginstructingllmseffectively} improves LLMs’ ability to acquire knowledge from raw documents through self-teaching.

Alternative approaches, such as GenRead~\cite{yu2022generate}, replace retrievers with LLM generators, using generated contexts to answer questions. Additionally, \citet{tang-etal-2024-b} propose the “A+B” generator-reader framework, facilitating new knowledge acquisition through CL.

\paragraph{Knowledge Conflict}
Evolving knowledge highlights conflicts between internal and external knowledge. Recent studies investigate the impact of knowledge conflicts on LLMs~\cite{ying-etal-2024-intuitive, xie2024adaptive, Marjanovi2024DYNAMICQA}. These studies find that such conflicts do affect LLM performance. For instance, \citet{ying-etal-2024-intuitive} find that LLMs tend to generate answers aligned with their internal knowledge, even when the provided external knowledge is correct.

%% file: chapters/data_collection.tex
\section{EvoWiki Dataset}
\label{sec:dataset_construction}

In this section, we outline the construction process of the EvoWiki dataset, which integrates 
several features, such as knowledge evolution levels, referenced context, multi-hop reasoning capabilities, and popularity attributes. We identify facts at various stages of evolution by comparing different temporal versions of English Wikidata\footnote{\href{https://www.wikidata.org/wiki}{https://www.wikidata.org/wiki}} (referred to as Wikidata). These facts are then anchored to English Wikipedia\footnote{\href{https://en.wikipedia.org/wiki}{https://en.wikipedia.org/wiki}} (referred to as Wikipedia) using distant supervision to ensure data integrity and provide referenced context. Additionally, we develop multi-hop reasoning questions based on the identified knowledge facts and incorporate extra attributes such as popularity.

\begin{figure}
    \setlength{\abovecaptionskip}{2pt}   
    \setlength{\belowcaptionskip}{0pt}
  \centering
  \includegraphics[width=0.48\textwidth]{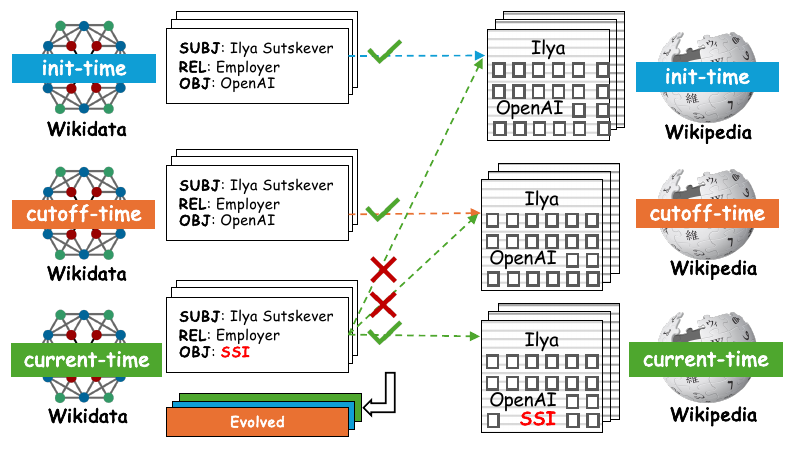}
  \caption{Evolution level identification process.}
  \label{fig:evo_level}
  \vspace{-3mm}
\end{figure}

\subsection{Knowledge Evolution 
Level Identification}
\label{subsec:knowledge_evolution_levels}
The evolution of a fact is determined in relation to the knowledge cut-off date of LLMs. Specifically, as shown in Figure \ref{fig:intro_fig}, facts are categorized into three levels: stable, evolved, and uncharted. Stable facts remain unchanged after the LLM's knowledge cut-off date. Evolved facts were established before the cut-off date but have undergone changes since. Uncharted facts represent entirely new information introduced after the cut-off date.
  
To determine the evaluation level of a fact, we introduce three key timestamps: \textit{init-time}, \textit{cutoff-time}, and \textit{current-time}. \textit{Init-time} represents an early point in time before which facts are well-established in LLMs, \textit{cutoff-time} is the knowledge cut-off date of the LLM, and \textit{current-time} is the time at which the evaluation is conducted. In our implementation, we set the \textit{init-time} to September 2021, the \textit{cutoff-time} to January 2024, and the \textit{current-time} to May 2024, aligning with the knowledge update timeline of popular LLMs, as detailed in the Appendix \ref{sec:cut-off-dates}. These timestamps are easily adjustable to accommodate different LLMs' knowledge update schedules, which enables the auto-update of the EvoWiki benchmark.

As shown in Figure \ref{fig:evo_level}, based on the three snapshots of Wikidata/Wikipedia, the evolution level of a fact is determined by analyzing changes across different timestamps. The classification rules are outlined as follows (detailed in Appendix \ref{sec:evolution_level}):

\begin{itemize}[leftmargin=*,nosep]
  \item \textbf{Stable}: facts that remain unchanged from \textit{init-time} to \textit{current-time}.
  \item \textbf{Evolved}: facts that are established before \textit{init-time} and exhibit changes between \textit{cutoff-time} (or \textit{init-time}) and \textit{current-time}.
  \item \textbf{Uncharted}: facts that are introduced after \textit{cutoff-time}.
\end{itemize}

Facts are categorized into distinct evolution levels. However, some of these facts may contain noise, such as unanswerable or inaccurate details. To mitigate this, we link each factual triple to its corresponding context on the relevant Wikipedia page using distant supervision, ensuring that the triple's value is referenced within that context.

\begin{table}[]
    \setlength{\abovecaptionskip}{5pt}   
    \setlength{\belowcaptionskip}{0pt}
  \centering
  \resizebox{\linewidth}{!}{
  \begin{tabular}{l|rrr}
  \toprule
      Data type & Num. of questions & Avg. length of context & Avg. popularity \\
      \midrule
      Stable  & 3,819 & 5,411.98 & 16,305.96  \\
      Evolved  & 3,491 & 4,451.90 & 42,807.55 \\
      Uncharted  & 2,954 & 5,014.30 & 24,039.57 \\
    
  \bottomrule
  \end{tabular}
  }
  
  \caption{Detailed Statistics of EvoWiki.}
  \label{tab:statics of evowiki}
  \vspace{-3mm}
\end{table}

\subsection{
Multi-dimensional Attributions}
\label{subsec:facts_refinement}

We further expand the dataset by incorporating additional attributes, including \textit{Referenced Context}, \textit{Multi-hop Reasoning}, and \textit{Popularity}. The overall statistics of the current version of the EvoWiki dataset are presented in Table \ref{tab:statics of evowiki}.

\paragraph{\textbf{Referenced Context}}
We restrict the entity type to humans and link the triples to their corresponding Wikipedia pages using the identical \textit{wiki\_link} of the entity. A fact is considered supported if the triple's object entity (or subject entity) is explicitly mentioned on the corresponding Wikipedia page of the subject entity (or object entity). For triples with multiple objects, we verify all objects and retain only those explicitly mentioned to ensure high quality. Additionally, for stable facts, the triples must be supported by the corresponding Wikipedia pages across all three timestamps. Evolved and uncharted facts must be supported by the \textit{current-time} version of the Wikipedia page but not by the previous version. This process ensures that the facts are answerable, accurate, and provide a reliable, high-quality context for each fact triple.
Based on distant supervision, we consider the short mentioned sentence as the golden context of the fact triple and the corresponding Wikipedia page as the golden document.

\paragraph{\textbf{Multi-hop Reasoning}}
Building on the refined fact triples and corresponding contexts, we further enhance the dataset by constructing multi-hop reasoning questions. To maintain high quality, we apply the same rigorous filtering process, retaining only those triples where the objects (or subjects) are explicitly mentioned in the corresponding context for each hop. To reduce ambiguity, triples in the middle hop are restricted to facts with single object. In our implementation, reasoning questions are extended up to three hops\footnote{We do not make a strict fine-grained distinction for hops in the main experiments, as the automated process might generate 3-hop questions with superficial reasoning, which degenerate into 2-hop questions.}.

To generate questions, we first use templates to create questions asking for the object entity of the triple in the last hop. For instance, given the triple \textit{(Barack Obama, spouse, Michelle Obama)}, a template question is ``\textit{Who is the spouse of Barack Obama?}''. Afterward, we employ GPT-4o-mini \cite{openai2024gpt4ocard} to refine the questions for improved naturalness. Prompts are provided in Appendix \ref{sec:prompts}. The answers correspond to the object entity labels of the last hop, with all objects considered correct for multi-object facts.

\paragraph{\textbf{Popularity}}
We also incorporate additional attributes, such as popularity, to enrich the dataset. Popularity is measured by the number of page views for the corresponding Wikipedia page. This metric provides insights into the relevance and significance of the facts, allowing for more comprehensive analysis and evaluation.

\begin{table}[]
    \setlength{\abovecaptionskip}{5pt}   
    \setlength{\belowcaptionskip}{0pt}
  \centering
  \resizebox{\linewidth}{!}{
  \begin{tabular}{l|ccc}
  \toprule
      \textbf{Metrics} & \textbf{Stable} & \textbf{Evolved} & \textbf{Uncharted}  \\
      \midrule
        Fluency &  99.17 / 95.69 & 94.58 / 95.56 & 95.00 / 95.42 \\
        Answerability &   96.67 / 94.44 & 94.17 / 95.69 & 92.92 / 92.64 \\
        Accuracy &   97.92 / 93.19 & 93.33 / 94.58 & 91.67 / 90.97 \\
    
  \bottomrule
  \end{tabular}
  }
  
  \caption{
  Human evaluation on data quality. The scores indicate the normalized average scores of single-hop questions (\%) / all questions (\%).}
  \label{tab:human_eval}
  \vspace{-3mm}
\end{table}

\subsection{Human Evaluation on Data Quality}
\label{subsec:quality_assurance}
To ensure data quality, we perform manual checks to validate the generated questions and answers. A human evaluation is carried out by four senior computational linguistics researchers on 180 randomly selected samples (20 samples for each hop level of each evolution type). The evaluation assesses each question-answer pair based on three criteria: fluency (whether the question is grammatically correct and flows smoothly), answerability (whether the question has clear and explicit answers), and accuracy (whether the provided answer is correct). 
The detailed annotation guidelines for the human annotators are presented in Appendix \ref{sec:annotation_guidlines}.
As shown in Table \ref{tab:human_eval}, all these three key aspects of data quality are verified by the human annotators. The evaluation results suggest that the questions are clear and easy to understand, as well as answerable, with the provided answers demonstrating high accuracy.
Annotators reported that potential inaccuracies in answers primarily stem from noise in Wikidata.

%% file: chapters/experiments.tex
\section{Experiments}
We evaluate two types of widely-adopted methods on the EvoWiki dataset: Retrieval-Augmented Generation (RAG) and Continual Learning (CL). In the RAG setting, models are required to retrieve relevant documents for the question from a knowledge source and generate answers based on the retrieved documents. In the CL setting, models are fine-tuned with newly introduced data. Additionally, we explore the performance of combining RAG and CL to assess potential improvements.

\begin{table*}[h]
    \setlength{\abovecaptionskip}{5pt}   
    \setlength{\belowcaptionskip}{0pt}
  \centering

  \begin{threeparttable}
  \resizebox{0.85\textwidth}{!}{
  \begin{tabular}{l|cc|cc|cc} 
  \toprule
  \multirow{2}{*}{\textbf{Method}} &
  \multicolumn{2}{c|}{\textbf{Stable}} & \multicolumn{2}{c|}{\textbf{Evolved}} & \multicolumn{2}{c}{\textbf{Uncharted}}\\
  \cline{2-3} \cline{4-5} \cline{6-7}
  & single-hop & multi-hop & single-hop & multi-hop & single-hop & multi-hop\\
  \midrule
  \multicolumn{7}{c}{\textbf{Meta-Llama-3.1-8B-Instruct}}\\
  \midrule
  Open-book & 86.87 & 56.40 & 75.24 (83.47) & 60.30 & 83.52 & 51.32 \\
  Closed-book & 31.61 & 22.17 & 6.96 (24.61) & 13.99 & 10.84 & 17.90 \\
  \midrule
  BM25 & 59.41 & 14.42 & 36.13 (53.78) & 13.85 & 44.93 & 15.47 \\
  Contriever & \textbf{77.90} & \textbf{19.37} & \textbf{48.99} (72.70) & \textbf{17.85} & \textbf{72.69} & \textbf{21.42} \\
  $\text{BM25}_\text{large corpus}$ & 51.77 & 14.81 & 28.12 (44.95) & 14.27 & 35.86 & 15.70 \\
  $\text{Contriever}_\text{large corpus}$ & 68.92 & 16.49 & 44.28 (67.99) & 14.41 & 64.85 & 18.72 \\
  \midrule
  CPT + Closed-book & 35.83 & 24.41 & 8.83 (28.12) & 15.85 & 15.07 & 20.38 \\
  SFT + Closed-book & 36.97 & 24.41 & 8.53 (28.12) & 17.34 & 15.15 & 20.59 \\
  CPT + SFT + Closed-book & 38.31 & 25.48 & 8.75 (29.32) & 17.85 & 15.86 & 20.98 \\
  SFT + CPT + Closed-book & \textbf{38.58} & \textbf{28.84} & \textbf{10.25} (31.19) & \textbf{18.22} & \textbf{17.27} & \textbf{22.41} \\

  \midrule
  CPT + Open-book & 87.94 & 59.06 & 70.98 (83.40) & 62.06 & 84.32 & 53.36 \\
  SFT + Open-book & \textbf{92.10} & 60.22 & \textbf{80.78} (88.56) & 62.90 & \textbf{89.34} & 55.07 \\
  CPT + SFT + Open-book & 90.69 & \textbf{60.27} & 79.66 (87.51) & \textbf{63.51} & 87.31 & 53.80 \\
  SFT + CPT + Open-book & 89.82 & 59.54 & 74.87 (85.71) & 63.27 & 86.52 & \textbf{55.34} \\

  \midrule
  CPT + Contriever & 77.70 & 22.73 & 44.05 (73.00) & 19.53 & 71.45 & 22.74 \\
  SFT + Contriever & \textbf{82.85} & 24.02 & \textbf{57.22} (79.36) & \textbf{20.22} & \textbf{78.85} & \textbf{24.84} \\
  CPT + SFT + Contriever & 79.64 & 24.19 & 49.74 (76.29) & 19.39 & 75.51 & 23.35 \\
  SFT + CPT + Contriever & 76.02 & \textbf{24.97} & 47.27 (74.05) & 20.18 & 73.13 & 23.40 \\

  \midrule
  \multicolumn{7}{c}{\textbf{Mistral-7B-Instruct-v0.3}}\\
  \midrule
  \noalign{\vskip 0.07cm} 

  Open-book & 87.68 & 60.57 & 77.56 (83.99) & 60.44 & 82.64 & 56.00 \\
  Closed-book & 29.81 & 23.12 & 5.83 (19.90) & 15.76 & 10.04 & 18.89 \\
  \midrule
  BM25 & 52.85 & 14.46 & 34.78 (50.49) & 16.08 & 44.14 & 16.46 \\
  Contriever & \textbf{73.14} & \textbf{22.17} & \textbf{52.43} (74.05) & \textbf{19.11} & \textbf{71.89} & \textbf{23.57} \\
  $\text{BM25}_\text{large corpus}$ & 40.32 & 14.25 & 26.33 (38.82) & 13.20 & 32.25 & 13.43 \\
  $\text{Contriever}_\text{large corpus}$ & 63.16 & 18.04 & 46.97 (67.02) & 15.20 & 61.85 & 20.04 \\
  \midrule
  CPT + Closed-book & 35.43 & 28.20 & 9.57 (28.57) & 18.83 & 14.98 & 23.57 \\
  SFT + Closed-book & \textbf{38.31} & \textbf{33.62} & \textbf{10.77} (30.29) & \textbf{21.62} & \textbf{16.30} & \textbf{27.53} \\
  \midrule
  CPT + Open-book & 88.61 & 60.27 & 78.53 (83.40) & 62.58 & 81.23 & 55.62 \\
  SFT + Open-book & \textbf{91.43} & \textbf{71.16} & \textbf{85.86} (89.75) & \textbf{73.18} & \textbf{89.07} & \textbf{66.19} \\
  \midrule
  CPT + Contriever & 74.28 & 26.43 & 52.88 (75.69) & 21.89 & 71.72 & 25.88 \\
  SFT + Contriever & \textbf{80.44} &\textbf{30.99} & \textbf{61.78} (78.98) & \textbf{24.27} & \textbf{76.04} &\textbf{29.29} \\
  \midrule
  
  \bottomrule
  
  \end{tabular}}
  \end{threeparttable}
  \caption{Main performance of the methods on EvoWiki. Values in parentheses indicate the precision of all answers that contain outdated answers.}
  \label{tab: direct_utilization}
  \vspace{-3mm}
\end{table*}

\subsection{Experimental Settings}

Our experiments are conducted using two widely used models: Llama-3.1-8B-Instruct (referred to as Llama) and Mistral-7B-Instruct (referred to as Mistral) on EvoWiki. 
The corpus is built from a 15K Wikipedia dump of golden documents, and provide an additional expanded version (denoted as \textit{large\_corpus}) that includes 370K randomly selected Wikipedia articles to simulate a more practical scenario. Each document is divided into 256-token chunks. The models answer questions in a zero-shot setting using a simple prompt (Appendix \ref{sec:prompts}). Performance is measured with the exact match (EM) metric, evaluating the percentage of questions answered correctly. For evolved data, we consider responses with the latest answer as correct and also compare results with outdated answers.

\textbf{Closed-Book and Open-Book QA.}
Closed-book and open-book QA represent the lower and upper performance bounds. In closed-book QA, models answer questions using their internal memory. In open-book QA, models are provided with a golden context, a concise yet informative sentence extracted from Wikipedia (Section \ref{subsec:facts_refinement}), ensuring minimal noise and optimal contextual support.

\textbf{RAG.}
We employ two retrieval models, BM25 \cite{10.1561/1500000019} and Contriever~\cite{izacard2022unsuperviseddenseinformationretrieval}, to fetch relevant documents. BM25, a sparse retrieval model, scores relevance using term frequency and inverse document frequency. Contriever, a dense retrieval model, encodes queries and documents into a shared embedding space, measuring relevance via cosine similarity. Models generate answers using the top-15 retrieved chunks as context.

\textbf{CL.}
We integrate new knowledge into the model using continual pre-training (CPT) and supervised fine-tuning (SFT). CPT trains the model on the corpus with a language modeling objective, while SFT fine-tunes the model on question-answer pairs generated by prompting Llama with the given context. Following \citet{jiang2024instructiontuned}, we also evaluate combinations of CPT and SFT. Implementation details are provided in Appendix \ref{sec:implementation-details}.

\begin{figure*}
    \setlength{\abovecaptionskip}{2pt}   
    \setlength{\belowcaptionskip}{0pt}
  \centering
  \includegraphics[width=\linewidth]{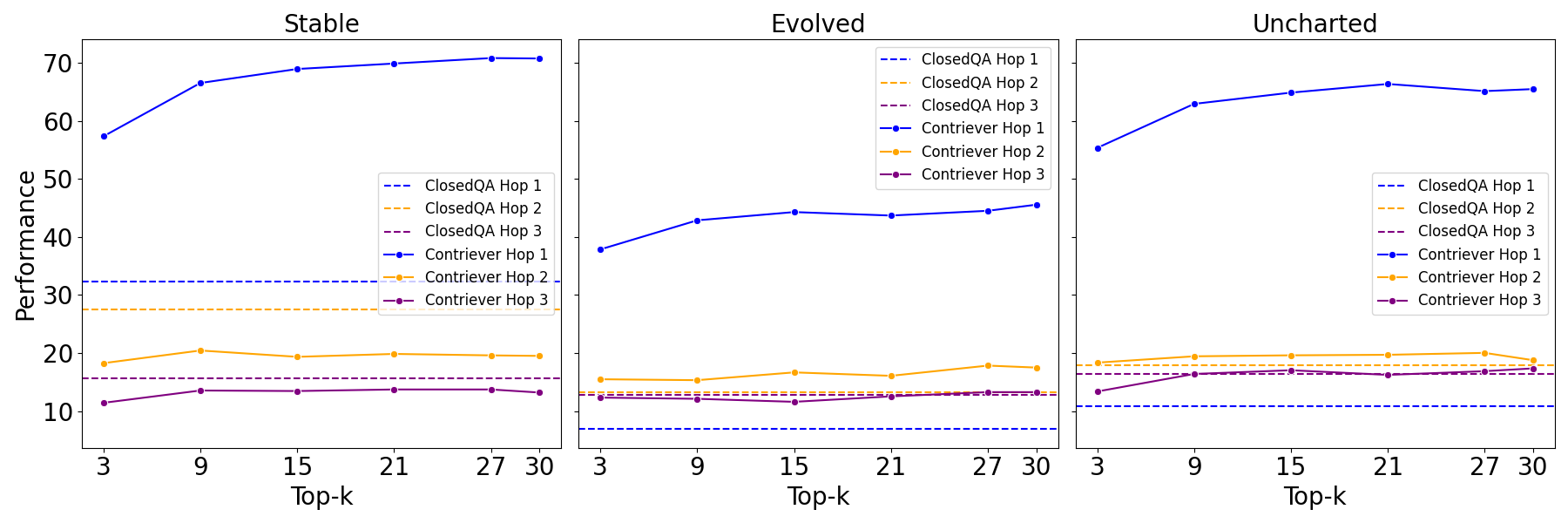}
  \caption{RAG performance across top-k values of Contriever; the dashed line represents closed-book QA results.}
  \label{fig:rag_num}
\end{figure*}

\begin{table*}[htb]\small
    \setlength{\abovecaptionskip}{5pt}   
    \setlength{\belowcaptionskip}{0pt}
  \centering

  \begin{threeparttable}
  \resizebox{0.95\textwidth}{!}{
  \begin{tabular}{l|cc|cc|cc} 
  \toprule
  \multirow{2}{*}{\textbf{Method}} &
  \multicolumn{2}{c|}{\textbf{Stable}} & \multicolumn{2}{c|}{\textbf{Evolved}} & \multicolumn{2}{c}{\textbf{Uncharted}}\\
  \cline{2-3} \cline{4-5} \cline{6-7}
  & Single-hop & Multi-hop & Single-hop & Multi-hop & Single-hop & Multi-hop \\
  \midrule
  \rowcolor[gray]{0.9}  
  Open-book & \textbf{86.87} & \textbf{56.40} & \textbf{75.24} (83.47) & \textbf{60.30 }& \textbf{83.52} & \textbf{51.32} \\
  SC Open-book | Memory & 64.84 & 28.32 & 53.78 (65.74) & 26.73 & 51.10 & 24.01 \\
  SC Open-book | Open-book  & 84.80 & 35.21 & 72.85 (81.53) & 42.68 & 80.53 & 36.56 \\
  \midrule
  \rowcolor[gray]{0.9}  
  BM25 & 59.41 & 14.42 & 36.13 (53.78) & 13.85 & 44.93 & 15.47 \\
  SC BM25 | Memory  & 50.84 & \textbf{16.19} & 28.12 (47.42) & 12.97 & 32.60 & 16.36 \\
  SC BM25 | BM25 & 58.20 & 11.88 & 36.13 (52.95) & 10.55 & 43.96 & 12.44 \\
  SC BM25 | Contriever & \textbf{72.94} & 15.80 & \textbf{47.57} (71.28) & \textbf{15.20} & \textbf{69.87} & \textbf{17.62} \\
  \midrule
  \rowcolor[gray]{0.9}  
  Contriever & \textbf{77.90} & \textbf{19.37} & \textbf{48.99} (72.70) & \textbf{17.85} & \textbf{72.69} & \textbf{21.42} \\
  SC Contriever | Memory  & 60.42 & 17.78 & 35.98 (58.41) & 14.41 & 44.05 & 17.02 \\
  SC Contriever | BM25 & 63.50 & 13.60 & 35.83 (55.05) & 12.04 & 46.52 & 13.93 \\
  SC Contriever | Contriever & 73.74 & 17.14 & 46.52 (70.83) & 15.34 & 69.07 & 17.84 \\
  
  \bottomrule
  \end{tabular}}
  \end{threeparttable}
  \caption{Performance of self-critique. `A | B' means using B as the reference context to check the answer of A. Values in parentheses indicate the precision of all answers
  that contain outdated answers.}
  \label{tab: self-critique}
  \vspace{-3mm}
\end{table*}

\subsection{Overall Results}

\textbf{Models perform better on stable facts than on evolved and uncharted facts.} As shown in Table \ref{tab: direct_utilization}, Both Llama and Mistral demonstrate expected results in the closed-book setting for single-hop questions, achieving an average of 31.61\% and 29.81\% on stable facts, 6.96\% and 5.83\% on evolved facts, and 10.84\% and 10.04\% for both models on uncharted facts. 
These results suggest models have reliable memory for knowledge they previously encountered but struggle to adapt to new knowledge relying solely on reasoning. Additionally, these findings validate the construction of EvoWiki.

With golden context, models perform well across all data types, though accuracy drops significantly on evolved facts. Performance on outdated answers matches that on other types of facts, suggesting conflicts between internal and external knowledge limit effective utilization.
Both RAG and CL improve performance across all data types but lag behind the open-book setting. Larger gaps for evolved and uncharted facts highlight the difficulty of integrating new knowledge into models.

\subsection{Retrieval-augmented Generation}

\textbf{RAG shows promising performance but struggles with multi-hop reasoning.}
With the use of RAG, the performance of both models on single-hop questions significantly improves, as shown in Table \ref{tab: direct_utilization}, with an increase of +27.80\%/46.29\% and +23.04\%/43.33\% on stable facts, +29.17\%/42.03\% and +28.95\%/46.60\% on evolved facts, and +34.09\%/61.85\% and +34.10\%/61.85\% on uncharted facts for BM25/Contriever, respectively.
However, performance on multi-hop questions is severely limited, with a noticeable degradation on stable and uncharted facts, even when compared to the closed-book setting.
Additionally, RAG experiences a performance drop when the corpus is enlarged. These results suggest that RAG’s effectiveness depends on the retrieval model’s ability to provide relevant information, which works well for simpler questions but introduces more noise than useful content when handling complex questions.

\textbf{RAG is influenced by noise, leading to negative effects on known knowledge.} To further explore the impact of noise, we conduct experiments with varying top-k retrieval settings, as shown in Figure \ref{fig:rag_num}. Increasing top-k improves performance initially, but beyond 15, the improvement flattens and even showing a downward trend. This trend is observed across all three types of data, suggesting that noise affects each evolution level similarly.

We also noticed that on the evolved and uncharted data, RAG's performance on multi-hop data exceeds that of the closed-book, while the opposite holds for stable data. Because of lacking of explicit keyword, the noise introduced in multi-hop retrieval is likely to be less relevant to the answer, and this noise do negatively affect the model's utilization of its known internal knowledge.

\textbf{Self-critique failed to improve the performance of RAG.}
Inspired by recent advancements in self-critique techniques~\cite{shinn2023reflexionlanguageagentsverbal, valmeekam2023largelanguagemodelsreally}, we investigated the potential of self-critique to enhance RAG by verifying the consistency between generated answers and contexts (or memory), enabling the model to revise its responses on their own. Experiments combining RAG with self-critique, as summarized in Table \ref{tab: self-critique}, revealed that self-critique did not improve RAG’s performance. While using stronger retrieval results as reference context enhanced weaker retrieval models, it still fell short of directly leveraging the stronger retrieval.
We attribute this limitation to that models tend to rely on their internal knowledge when faced with uninformative context. Distinguishing when to rely on internal memory versus retrieved context remains a non-trivial challenge.

\subsection{Continual Learning}
\textbf{CL shows modest yet consistent improvement.}
In Table \ref{tab: direct_utilization}, on single-hop questions, both CPT and SFT yield notable gains, with +4.22\%/5.36\% and +5.62\%/8.50\% on stable facts, and +4.23\%/4.31\% and +4.94\%/6.26\% on uncharted facts for Llama and Mistral, respectively. On evolved fact, when only considering the latest answer, improvements are smaller, at +1.87\%/1.57\% and +3.74\%/4.94\% for Llama and Mistral. Including outdated answers brings performance closer to stable and uncharted fact, highlighting challenges in modifying knowledge.
Unlike RAG, CL does not negatively impact multi-hop questions but instead improves performance, demonstrating its potential in integrating knowledge without sacrificing multi-hop scenarios.

\textbf{CPT and SFT are complementary.} We further explore the performance of combining CPT and SFT. Drawing inspiration from~\cite{jiang2024instructiontuned}, we evaluate the impact of different training orders of CPT and SFT. As shown in Table \ref{tab: direct_utilization}, in closed-book QA, improvements are observed across all data types when combining CPT and SFT, with the best performance achieved when applying SFT first, followed by CPT—consistent with the findings in~\cite{jiang2024instructiontuned}. These results suggest a synergistic effect between CPT and SFT in integrating new knowledge into the model.

\begin{figure}
\setlength{\abovecaptionskip}{2pt}   
\setlength{\belowcaptionskip}{0pt}
  \centering
  \includegraphics[width=1\linewidth]{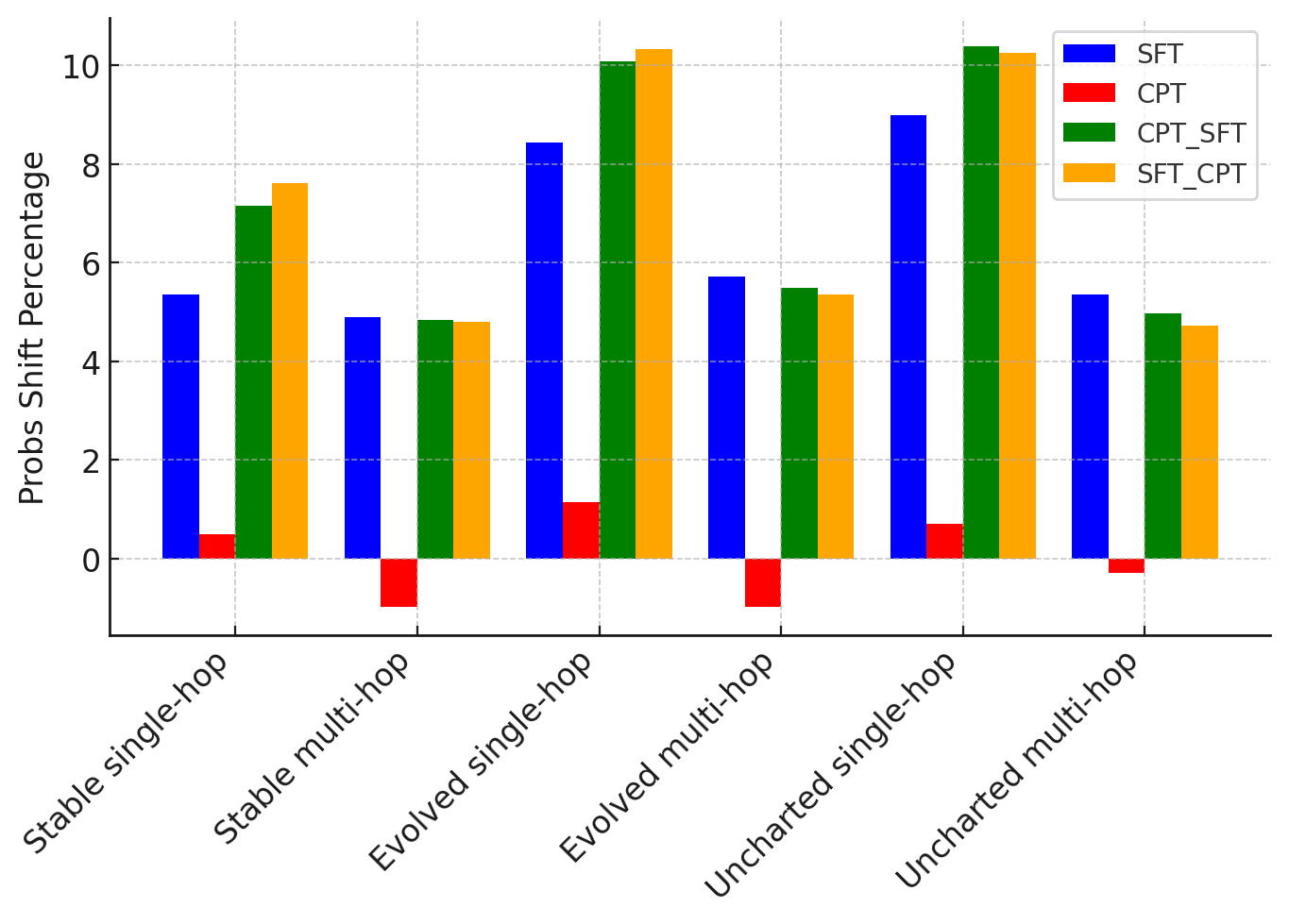}
  \caption{Probability shift (\%) of CL methods on Llama for the first token of the golden answer.}
  \label{fig:prob_shift}
  \vspace{-3mm}
\end{figure}

\textbf{SFT demonstrates superior knowledge integration over CPT.} It is non-trivial to compare CPT and SFT using the EM metric, as their performance is quite similar. Therefore, we introduce a simplified Persuasion Score~\cite{du2024contextversuspriorknowledge} that measures how the CL method affects the model's probability of generating the correct answer. As shown in Figure \ref{fig:prob_shift}, the probability shifts reveal that SFT is much better at correcting the model's predictions than CPT. Furthermore, the combination of CPT and SFT shows a significant impact regardless of the order in which they are applied.

\begin{figure}[t]
\setlength{\abovecaptionskip}{2pt}   
\setlength{\belowcaptionskip}{0pt}
    \centering
    \includegraphics[width=\linewidth]{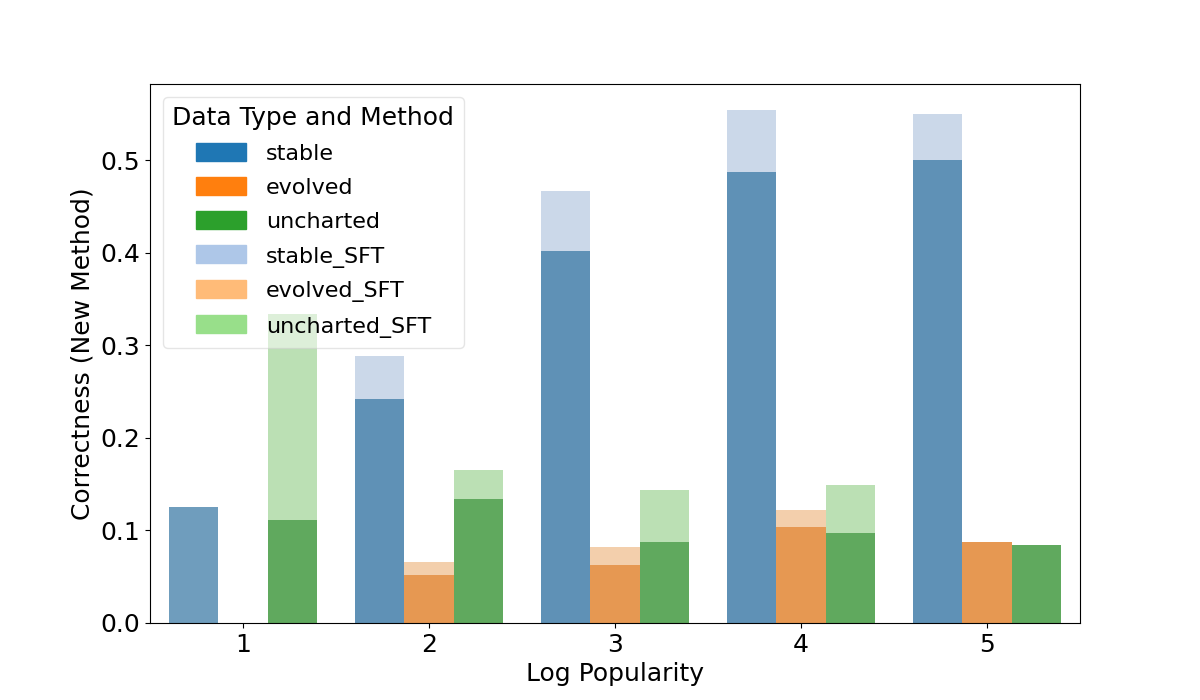}
  \caption{Popularity effects of SFT on Llama. Due to data scarcity, we aggregated the popularity levels of 0 and 1 into a single category, as well as levels 5 and 6.}
  \label{fig:popularity_effects}
  \vspace{-3mm}
\end{figure}

\textbf{Popularity influences the effectiveness of CL.}
Popularity is a well-known factor that affects the performance of knowledge acquisition~\cite{mallen-etal-2023-trust}. To examine this, we follow recent research that considers Wikipedia page views as a measure of popularity and investigate its influence across different levels of knowledge evolution.

As illustrated in Figure \ref{fig:popularity_effects}, the results show different trends based on the data's evolution level. In the closed-book QA setting, stable data exhibits a positive correlation with popularity, which is intuitive since more popular knowledge is likely to have been encountered by the model. In contrast, both evolved and uncharted data show minor correlation with popularity, indicating that the model lacks relevant knowledge.

When augmented with SFT, stable data continues to show a positive correlation with popularity, while evolved data highlights the difficulty of reflecting changes in the model's internal knowledge. Interestingly, the model appears to learn new knowledge more effectively when the popularity is lower. For example, the improvement is significantly greater when the log popularity is 1 compared to when it is 5. These findings suggest that, rather than merely increasing the data scale, the proportion of training data should account for the popularity of the knowledge being learned.

\subsection{Combination of RAG and CL}

RAG shows strong performance on single-hop questions but is limited on multi-hop questions, while CL demonstrates modest yet consistent improvement on both single-hop and multi-hop questions. A natural approach is to combine RAG and CL to leverage the strengths of both methods. Thus, we conducted experiments with different combinations of RAG and CL, as shown in Table \ref{tab: direct_utilization}.

\textbf{The combination of RAG and CL demonstrates a synergistic effect.}
Integrating RAG with CL enhances performance across data types, particularly on multi-hop questions, compared to RAG with an untuned model. By updating internal knowledge through CL, the model provides more accurate answers when confronted with uninformative context from the retriever. This highlights the potential of combining both methods to leverage complementary strengths effectively.

%% file: chapters/appendix.tex

\section{Cut-off Dates of LLMs}
\label{sec:cut-off-dates}
According to the model cards of the LLMs, we statically collected the cut-off date of the LLMs as shown in below.
\begin{itemize}
    \item chatGPT-4: Up to December 2023.
    \item chatGPT-3.5: Up to September 2021.
    \item Llama3: March 2023 for the 7B and December 2023 for the 70B.
    \item Llama2: Between January 2023 and July 2023.
    \item Llama1: Between December 2022 and February 2023.
    \item Vicuna 1.1: Between March 2023 and April 2023.
    \item Mistral: No official cut-off date.
\end{itemize}

\section{Detail of Evolution Level Identification}
\label{sec:evolution_level}

Identify the evolution level of a fact, primarily based on the changes in Wikidata and Wikipedia at three different time snapshots. As shown in the Figure \ref{fig:evo_level}, we first determine the same triples across the three snapshots based on the unique identifier of the fact triple in Wikidata. Then we determine whether the triple has changed at cutoff-time or current-time; if not, it is temporarily marked as stable data; otherwise, it is considered evolved data. Next, we look for facts in the current-time Wikidata data that did not appear in init-time and cutoff-time, and these facts are temporarily marked as uncharted.   

Next, to further ensure data quality, we added a distant supervision process to ensure consistency across Wikipedia. Our strategy is as follows: for Stable facts, we ensure that the corresponding fact mentions can be found in all three Wikipedia snapshots. For Evolved facts, the fact before the change should have a mention in the corresponding Wikipedia, while the fact after the change should only be mentioned in the Wikipedia snapshot from the time the change occurred and not in earlier snapshots.  For uncharted facts, the mention should only exist in the current-time Wikipedia snapshot.

\section{Implementation Details of Continual Learning}
\label{sec:implementation-details}
For continual pre-tranining, we simply fine-tune the model with the 15K Wikipedia docuemnts with a language modeling objective. We train the model in 3 epochs with a batch size of 4, using Adam~\cite{kingma2014adam} optimizer with learning rate of 5e-6, and a maximum sequence length of 2048. We use the same hyperparameters for all models.

For supervised fine-tuning, we first generate the SFT data with Meta-Llama-3.1-8B-Instruct. Each document of Wikipedia are splited into multiple chunks with a maximum 512 tokens. Then we prompt the model to generate 6 questions for each chunk. We finally get 552K question-answer pairs as the SFT data. We fine-tune the model with the SFT data for 3 epochs with a batch size of 32, using Adam optimizer with learning rate of 5e-6, and a maximum sequence length of 256. We use the same hyperparameters for all models.

All implementations are conducted on 4 Nvidia A6000 GPUs. We use the Huggingface's transformers library~\cite{wolf2020huggingfacestransformersstateoftheartnatural}, and implementate parameter-efficient fine-tuning with Lora~\cite{hu2021loralowrankadaptationlarge} and set rank 16 and alpha 256.

\onecolumn

\section{Human Evaluation Guidelines}
\label{sec:annotation_guidlines}
The human evaluation guidelines for data quality validation are presented in Table \ref{tab: guideline}.

\begin{table*}[ht]
\small
\renewcommand{\arraystretch}{1.5}
\centering
\begin{tabular}{|p{3cm}|p{12cm}|}
\hline
\rowcolor[gray]{0.9}  
\multicolumn{2}{|c|}{\textbf{Guideline of Data Quality Evaluation}} \\
\hline
\multicolumn{2}{|p{15cm}|}{This evaluation focuses on the \textit{Fluency}, \textit{Answerability}, and \textit{Accuracy} of the generated question-answer pairs. Each question will have referenced context, referenced document, and two corresponding answers: the latest answer and all answers (where the latest answer and all answers are the same except for the evolved data). Accuracy is evaluated based on the latest answer.} \\
\hline
\rowcolor[gray]{0.9}  
\multicolumn{2}{|c|}{\textbf{Case}} \\
\hline
\textbf{Question:} & What is the occupation of Ashley Neal? \\
\hline
\textbf{Latest Answer:} & ['driving instructor', 'YouTuber'] \\
\hline
\textbf{All Answer:} & ['driving instructor', 'YouTuber', 'association football player'] \\
\hline
\textbf{Referenced Context} & ['Retired from football, Neal now works as a driving instructor and YouTuber.', 'He is now a driving instructor and instructor trainer.']\\
\hline
\textbf{Referenced Document} & ['Ashley Neal (born 16 December 1974) is an English former professional footballer who played as a defender ...  as of 16th December 2023 it had over 5,700 subscribers.'] \\
\hline
\rowcolor[gray]{0.9}  

\hline 
\rowcolor[gray]{0.9}  
\multicolumn{2}{|c|}{\textbf{Scoring Guide}} \\

\hline
\multirow{3}{*}{\textbf{Fluency}} & \textbf{3:} The question is perfectly clear and grammatically correct, with no ambiguities or errors. \\
\cline{2-2}
& \textbf{2:} The question is mostly clear but contains minor grammatical errors or slight ambiguities that do not hinder understanding. \\
\cline{2-2}
& \textbf{1:} The question is unclear, incomplete, or contains major grammatical errors that make it difficult to understand. \\
\hline
\multirow{3}{*}{\textbf{Answerability}} & \textbf{3:} The question is highly specific and can be answered unambiguously based on the provided context. \\
\cline{2-2}
& \textbf{2:} The question is somewhat specific but may lead to multiple interpretations or require additional clarification. \\
\cline{2-2}
& \textbf{1:} The question is vague or too broad, making it difficult to determine an exact answer. \\
\hline
\multirow{3}{*}{\textbf{Accuracy}} & \textbf{3:} The provided answer completely and accurately addresses the question without any inconsistencies. \\
\cline{2-2}
& \textbf{2:} The provided answer addresses the question partially, with minor inaccuracies or missing details. \\
\cline{2-2}
& \textbf{1:} The provided answer does not accurately address the question or is irrelevant to the question. \\
\hline
\end{tabular}
\caption{Human evaluation guidelines for data quality validation.}
\label{tab: guideline}
\end{table*}

\section{Prompts}
\label{sec:prompts}

\subsection{Question Generation}
The following prompt is used for question generation. The placeholders inside the single curly braces will be replaced respectively with the corresponding number of hops, triple strings, answer lists, and template questions.

\begin{quote}\small
    \begin{tcolorbox}[size=title,opacityfill=0.1]
        This is a \{hop\_num\}-hop question generation task. You are given \{hop\_num\} factual triples. Each triple consists of a subject entity, a relation, and an object entity. You should generate a question that ask about the last hop object entity. For a given triple, you should first understand the factual triples about what the fact is about. Then you need to union the relations of the multiple hops to generate a question that can be answered by the answer list.

        The question should follow the below requirements:

        - The question could only mention the subject entity of the first hop and the relations of the multiple hops. DO NOT mention any other entities.

        - The question should be generated based on the union of the relations of the multiple hops.

        - The question should be a valid question that can be answered by the answer list.

        - You are given a template question. You should rewrite the template question to make it natural. DO NOT introduce any new information that is not in the template question.
        
        For example, you are given the triples to generate a 2-hop question:

        hop1: [Ksenija Zadorina](Q457910), [country of citizenship](P27), [[Russia]]([Q159])

        hop2: [Russia](Q159), [follows](P155), [[Soviet Union]]([Q2164])

        answer list: [Soviet Union]

        template question: What is the follows of the country of citizenship of Ksenija Zadorina?

        Understanding the factual triples:

        This is a 2-hop relation. The first hop can be interpreted as: “Ksenija Zadorina has the country of citizenship as Russia.” This means that Ksenija Zadorina is a Russian citizen. The second triple can be interpreted as: “Russia follows the Soviet Union.” This likely refers to the historical transition where Russia is considered the successor state to the Soviet Union.

        Based on these triples, I can generate a 2-hop question by rewriting the template question to make it natural: Which entity does the country of citizenship of Ksenija Zadorina follow? And the answer is [Soviet Union], which is aligned to the requirement that the answer should be in the answer list. In this question, only mentioned the subject entity of the first hop and the relations of the multiple hops. The question is a valid question that can be answered by the answer list.

        Quetion: Which entity does the country of citizenship of Ksenija Zadorina follow?

        Answer: Soviet Union
        
        Now, you are given the following triples to generate a \{hop\_num\}-hop question:

        \{triple\_str\}

        answer list: \{answer\_list\}

        template\_question: \{template\_question\}

        Understanding the factual triples:
    \end{tcolorbox}
    \end{quote}

\subsection{SFT Data Generation}

The following prompt is used for generating SFT data. The placeholders inside the single curly braces will be replaced with the Wikipeida title and dump context.

\begin{quote}\small
    \begin{tcolorbox}[size=title,opacityfill=0.1]
I want you to act as a question writer expert. Your objective is to write **10** really complex and difficult question according to the given context make those famous AI systems (e.g., ChaGPT and GPT4) a bit harder to handle.

\#\# Generate Criterion\\
1. The question should be answerable without the given context. The question descirption should contain as much background information as possible, so the LLM can understand what the question is asking and where to find the answer.\\
2. The question should require llm to have already learnt and understood the context carefully so they can directly give the answer.\\
3. Ensure that you can confidently answer the questions you are proposing, if you can not answer it correctly or have no related knowledge about the entity please return "None".\\
4. Provide the only one correct answer to the generated question\\
5. The output format is as follows:\\
Question-Answer 1:\\
Question: \{\{the first generated question according to the fact and the context\}\}\\
Answer: \{\{the correct answer\}\}\\
Question-Answer 2:\\
Question: \{\{the second generated question according to the fact and the context\}\}\\
Answer: \{\{the correct answer\}\}\\
...

\#\# Title\\
\{title\}\\

\#\# Context\\
\{context\}\\

\#\# Response\\
Question-Answer 1:
    \end{tcolorbox}
\end{quote}

\subsection{Answer Without Context}
The following prompt is used for performing closed-book QA. The placeholders inside the single curly braces will be replaced with questions in the dataset.

\begin{quote}\small
    \begin{tcolorbox}[size=title,opacityfill=0.1]
        Answer the question directly with a single word or short phrase representing the most recent answer.

        The response format is as follows:

        \# Answer

        The correct answer: {{your answer}}

        \# Question

        \{question\}

        \# Answer

        The correct answer:
    \end{tcolorbox}
\end{quote}

\subsection{Answer With Context}

The following prompt is used for performing open-book QA and RAG. The placeholders inside the single curly braces will be replaced with questions and referenced context (or retrieved chunks).

\begin{quote}\small
    \begin{tcolorbox}[size=title,opacityfill=0.1]
        Answer the question directly based on the latest context, using a single word or short phrase.

        The response format is as follows:

        \# Answer

        The correct answer: {{your answer}}

        \# Context

        \{context\}

        \# Question

        \{question\}

        \# Answer
        The correct answer:
    \end{tcolorbox}
\end{quote}

\subsection{Self-Critique Prompt}

The following prompt is used for performing self-critique. The placeholders inside the single curly braces will be replaced with questions and the answer to be judged.

\begin{quote}\small
    \begin{tcolorbox}[size=title,opacityfill=0.1]
        Check if the student answer of the question is correct, answer with Yes/No, and provide the correct answer if it's not correct.

        The response format is as follows:

        \# Answer

        Yes/No: {{your reason}}

        The correct answer: {{your answer}}

        For example, if the studen answer is correct, your response is:

        \# Answer

        Yes: The student answer is correct 

        The correct answer: {{studen answer}}

        If the student answer is not correct, your response is:

        \# Answer

        No: The correct answer is {{correct answer}} which is {{reason}}

        The correct answer: {{correct answer}}

        Now, check the student answer below:

        \# Question

        \{question\}

        \# Student Answer

        \{first\_answer\}

        \# Answer
    \end{tcolorbox}
\end{quote}